\documentclass[letterpaper, 10 pt, conference]{ieeeconf}  % Comment this line out if you need a4paper

\IEEEoverridecommandlockouts                              % This command is only needed if 
\usepackage{color}
\usepackage{amsmath}
\usepackage{amsfonts}
\usepackage{amssymb}
\usepackage{graphicx}
\PassOptionsToPackage{hyphens}{url}\usepackage{hyperref}

\title{\LARGE \bf
  Efficient Online Transfer Learning for 3D Object Classification\\ in Autonomous Driving
}

\author{Rui Yang$^1$, \emph{Student Member, IEEE}, Zhi Yan$^{1*}$, \emph{Senior Member, IEEE},\\ Tao Yang$^2$, Yassine Ruichek$^1$, \emph{Senior Member, IEEE}% <-this % stops a space
  \thanks{The work is supported by the China Scholarship Council (CSC-UT-INSA program), the Contrat de Plan \'Etat-R\'egion (CPER 2015-2020 Mobilitech), and the Fundamental Research Funds for the Central Universities (NPU, No. 31020200QD045)}% <-this % stops a space
  \thanks{$^1$CIAD UMR 7533, Universit\'e Bourgogne Franche-Comt\'e, UTBM, F-90010 Belfort, France.
    {\tt\small firstname.lastname@utbm.fr}}%
  \thanks{$^2$Unmanned System Research Institute, Northwestern Polytechnical University, China. {\tt\small yangtao@nwpu.edu.cn}}%
  \thanks{$^{*}$Corresponding Author: {\tt\small zhi.yan@utbm.fr}}
}

\begin{document}

\maketitle
\thispagestyle{empty}
\pagestyle{empty}

%%%%%%%%%%%%%%%%%%%%%%%%%%%%%%%%%%%%%%%%%%%%%%%%%%
\begin{abstract}
%%%%%%%%%%%%%%%%%%%%%%%%%%%%%%%%%%%%%%%%%%%%%%%%%%
%———————————————— Version - 1.0 —————————————————%
Autonomous driving has achieved rapid development over the last few decades, including the machine perception as an important issue of it.
Although object detection based on conventional cameras has achieved remarkable results in 2D/3D, non-visual sensors such as 3D LiDAR still have incomparable advantages in the accuracy of object position detection.
However, the challenge also exists with the difficulty in properly interpreting point cloud generated by LiDAR.
This paper presents a multi-modal-based online learning system for 3D LiDAR-based object classification in urban environments, including cars, cyclists and pedestrians.
The proposed system aims to effectively transfer the mature detection capabilities based on visual sensors to the new model learning based on non-visual sensors through a multi-target tracker (i.e. using one sensor to train another).
In particular, it integrates the Online Random Forests (ORF)~\cite{saffari2009line} method, which inherently has the abilities of fast and multi-class learning.
Through experiments, we show that our system is capable of learning a high-performance model for LiDAR-based 3D object classification on-the-fly,
which is especially suitable for robotics in-situ deployment while responding to the widespread challenge of insufficient detector generalization capabilities.
\end{abstract}

%%%%%%%%%%%%%%%%%%%%%%%%%%%%%%%%%%%%%%%%%%%%%%%%%%
\section{Introduction}
\label{sec:introduction}
%%%%%%%%%%%%%%%%%%%%%%%%%%%%%%%%%%%%%%%%%%%%%%%%%%
%———————————————— Version - 1.0 —————————————————%
Object detection in 3D space is an essential portion for autonomous driving, which contains object localization and classification tasks.
The main sensors currently used for this purpose include cameras~\cite{chen2016monocular,mousavian20173d,xu2018multi} and LiDARs~\cite{chen2017multi,zhou2018voxelnet,kidono11iv}, while the latter is considered to be a strong support for the former, due to its high precision of range measurement and less sensitive to light conditions.
The well-known case can be traced back to the 2007 DARPA Grand Challenge, in which the Stanford Junior robot car's primary sensor for obstacle detection such as pedestrians, signposts, and cars, is a 64-layer 3D LiDAR~\cite{montemerlo08jfr}.
However, although highly anticipated, it is not straightforward to identify objects in the point cloud generated from such sensors, because some high recognition features such as texture and color are missing, making false positives more likely.

\begin{figure}[t]
  \centering
  \includegraphics[width=\columnwidth]{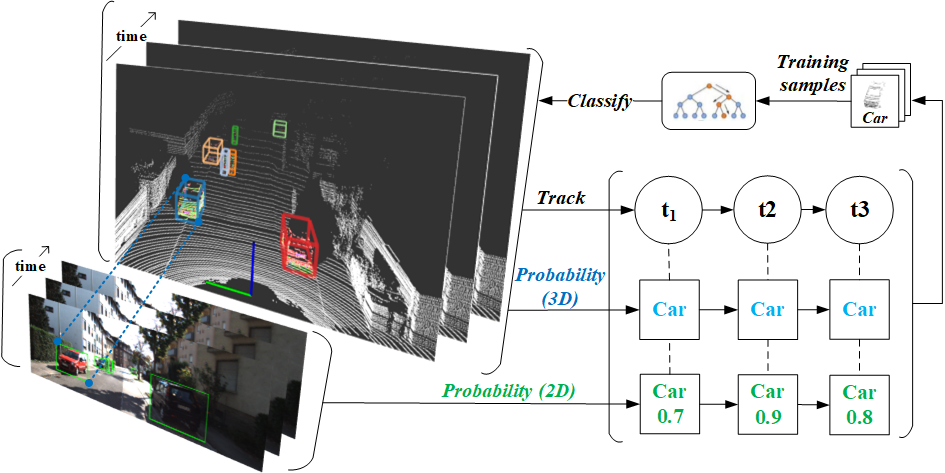}
  \caption{Conceptual diagram of the proposed system. First, the output (2D bounding boxes shown in the lower left) of a high-performance 2D image object detector is used for annotation of 3D point cloud data (3D bounding boxes shown in the upper left). Then a multi-target tracker is used to correlate the clusters in the point cloud to generate a series of learning examples. Finally, a classifier is trained on-the-fly by using the Online Random Forests (ORF)~\cite{saffari2009line}.}
  \label{fig:framework}
\end{figure}

More challenging, in response to the changing environment~\cite{yz20iros,tom17tro} and environment changes~\cite{zhimon20jist,ls18icra}, a feature-based model usually needs to be fine-tuned or even retrained to achieve acceptable performance over time or in new scenarios.
An approach to eliminate this dilemma is to train a generalized model.
However, the latter requires a large amount of labelled data, usually associated with high labor and machine costs.
As for 3D LiDAR data, labelling sparse point clouds is tedious work and prone to human error, in particular if many variations of object pose, shape and size need to be correctly classified.
Furthermore, timely learning of new samples to keep the model up-to-date is highly desirable to ensure the object detection performance of autonomous driving.

Compared with traditional offline learning (i.e. statistics), online learning is a paradigm more in line with the way humans learn.
It usually only requires a few~\cite{yz19auro,teichman11rss} or even no~\cite{yz18iros} manually annotated samples to learn a high-performance model.
Apart from this, the benefits it brings also include alleviating the computational burden when updating the model and improving the timeliness of the model, which is fully consistent with the development needs of autonomous driving for in-situ deployment~\cite{ls18icra} and long-term autonomy~\cite{yz20iros,tom17tro,Kunze2018}.

Based on today's popular multi-modal perception solutions, in this paper, we propose an efficient online transfer learning pipeline for 3D LiDAR-based object classification in autonomous driving.
An intuitive graphical description is shown in Fig.~\ref{fig:framework}.
First, we extract object information from the 2D images, including the label associated with probability and the 2D bounding box.
Meanwhile, the point cloud generated by the 3D LiDAR is segmented into clusters, and the 2D image-based object information is associated with the clusters via the multi-sensor calibration and the synchronization of the sensory data.
In other words, an image-based detector is used to automatically annotate the point cloud data to realize the knowledge transfer.
Then, the clusters in different frames are correlated by a multi-target tracker, the probability of which type of object each track belongs to is calculated, and the high-confidence track is input into the Online Random Forests (ORF)~\cite{saffari2009line} module as training samples.
It is worth pointing out that, $1)$ the random forest (RF) is inherently fast and suitable for multi-class model learning, and $2)$ the efficiency of learning is reflected in the fact that only a part of the clusters in the track are labelled due to the limited visual scope of the camera, but the track usually contains a large number of samples for learning due to the large-range and long-distance sensing capabilities of the 3D LiDAR.

The contributions of this paper are twofold.
First, we propose a novel implementation of the multi-sensor-based online learning framework~\cite{yz18iros} to allow efficient migration of mature image-based detection capabilities to 3D LiDAR, and use ORF to achieve rapid incremental learning.
Second, for the first time, we apply our online learning framework to 3D LiDAR-based object classification including cars, cyclists and pedestrians, in urban environments for autonomous driving, and open source our code based on ROS~\cite{ros} and Autoware~\cite{autoware} to the community: \url{https://github.com/epan-utbm/efficient_online_learning}.

%%%%%%%%%%%%%%%%%%%%%%%%%%%%%%%%%%%%%%%%%%%%%%%%%%
\section{Related Work}
\label{sec:related_work}
%%%%%%%%%%%%%%%%%%%%%%%%%%%%%%%%%%%%%%%%%%%%%%%%%%
%———————————————— Version - 0.1 —————————————————%
Online machine learning, with other closely related paradigm such as lifelong, transfer, continual, open-ended, and self-supervised learning, to name but a few, mainly refers to constantly update the knowledge model over time, and typically without human intervention.
Pioneering work in this field can be traced back more than two decades~\cite{thrun94iros}, in which a lifelong learning perspective for mobile robot control is presented.
With the rapid development of various related technologies including hardware and algorithms, in recent years, research on robotic online learning has become more and more extensive~\cite{yz19auro,yz17iros,teichman11rss,yz18iros,majer19ecmr,broughton20ras}.
In particular, Teichman and Thrun~\cite{teichman11rss} presented a semi-supervised learning to the problem of track classification in 3D LiDAR data based on Expectation Maximization (EM) algorithm, which illustrated that learning of dynamic objects can benefit from tracking system.
The proposed learning procedure starts with a small set of hand-labelled seed tracks and a large set of background tracks, while the latter is pre-collected in areas without any pedestrians, bicyclists, or cars.
In contrast, our research further focuses on in-situ learning in the absence or with incomplete background knowledge.

On the other hand, perception systems based on multi-modal sensors are still the preferred solution for self-driving car developers, as there is no almighty and perfect sensor so far, and they all have limitations and edge cases~\cite{yz20iros}.
In the past ten years, extensive research has been conducted on the use of multiple sensors for 3D object detection and tracking in the field of autonomous driving, and it is expected that more competitive performance can be obtained by effectively integrating the advantages of different sensors~\cite{chen2017multi,wang2018fusing}.
As one of the main sensors, the camera is favored by academia and industry due to its relatively low cost and capacity of providing rich semantics about the environment. 
And with the rapid development of deep learning technology, image-based detection have shown convincing results~\cite{chen2016monocular,mousavian20173d,xu2018multi,li2020rtm3d}.
However, the performance of methods based on such passive sensors are subject to lighting conditions and the relatively slow detection speed, and it is difficult to provide accurate 3D position information of the object.

Although with relatively high cost, 3D LiDARs have been widely used for autonomous driving as they are capable of providing large-scale, long-distance and high-precision distance measurement without considering the lighting conditions.
For example, Zhou and Tuzel~\cite{zhou2018voxelnet} proposed an end-to-end trainable deep network architecture that transforms LiDAR point cloud data into dense tensors (called voxel) to enable GPU-accelerated computing.
Alejandro et al.~\cite{gonzalez2016board} explored the fusion of RGB and depth maps obtained by high-definition LiDAR detection and ranging in the multi-modal component using RF as a local expert to boost the accuracy.
Nevertheless, these methods relies on considerable amount of manually annotated training data thus offline learned models to achieve effective object detection.

Through investigation of the state-of-the-art, none of the existing work in autonomous driving exploits information for multi-class learning from multi-sensor-based tracking in urban environments that our proposal does, that is, using one sensor to train another on-the-fly by fusing detections from both.
Our work actually combines the advantages of multi-modal system diversity with the efficiency of semi-supervised learning, and integrates them into an online framework.

%%%%%%%%%%%%%%%%%%%%%%%%%%%%%%%%%%%%%%%%%%%%%%%%%%
\section{General Framework}
\label{sec:general_framework}
\begin{figure}[t]
  \centering
  \includegraphics[width=\columnwidth]{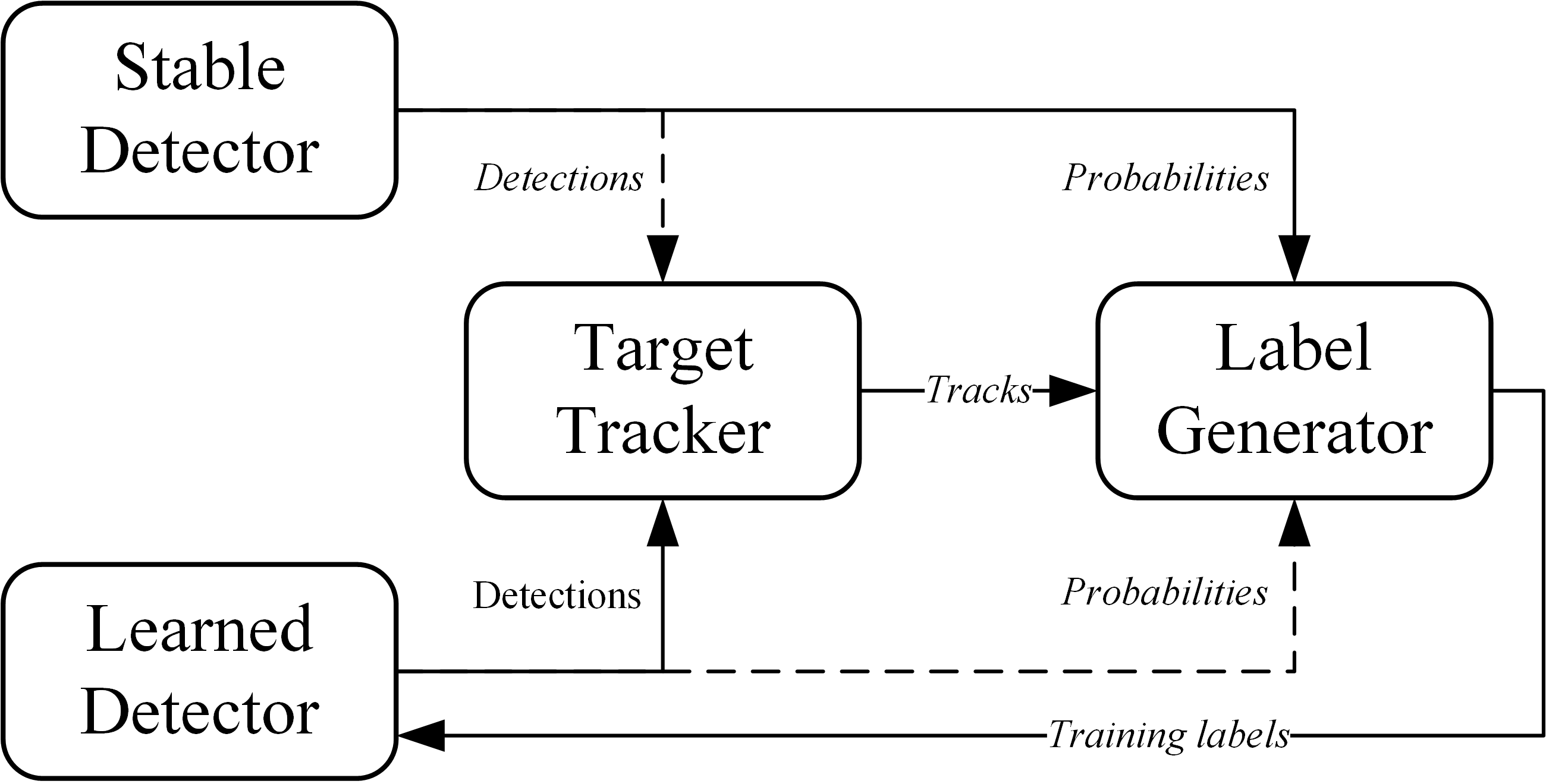}
  \caption{Block diagram of the general framework.}
  \label{fig:general_framework}
\end{figure}

%%%%%%%%%%%%%%%%%%%%%%%%%%%%%%%%%%%%%%%%%%%%%%%%%%
%———————————————— Version - 0.1 —————————————————%
In this section, we recall the general framework (illustrated in Fig.~\ref{fig:general_framework}) of online transfer learning that was previously proposed in~\cite{yz18iros}, which consists of four components: the \emph{Stable Detector} and the \emph{Learned Detector} are considered as the data entry, the \emph{Target Tracker} synthesizes the detections for multiple target tracking, and the \emph{Label Generator} provides labelled training data for online learning of the classifier within the \emph{Learned Detector}.

\subsection{Stable Detector and Learned Detector}
%———————————————— Version - 0.1 —————————————————%
We assume two types of detectors in our general framework.
The \emph{Stable Detector} has good detection ability, which is typically pre-trained and provides a high level of confidence.
The \emph{Learned Detector} has no detection capability at the beginning, but it can continuously learn with the help of the \emph{Stable Detector} over time.
It is hoped that under this framework, the detection capabilities of the \emph{Stable Detector} can be transferred to the \emph{Learned Detector} on-the-fly.
This mechanism is especially designed for those sensors whose data are difficult to annotate or features extracted from the data change greatly with the environment.
From a machine learning perspective, the \emph{Stable Detector} can provide labelled data, while the \emph{Learned Detector} can provide both unlabelled and labelled data (as it is constantly learning).
This shows that the evolution of the \emph{Learned Detector} can be further regarded as semi-supervised learning to some extent.

\subsection{Target Tracker and Label Generator}
%———————————————— Version - 0.1 —————————————————%
The \emph{Target Tracker} plays a critical role in our framework.
It correlates the detections from different detectors thereby improving the performance of object tracking on the one hand, and establishes a pipeline between the stable and the learned detectors thereby boosting the learning of the latter on the other hand.
The \emph{Label Generator} obtains tracks generated by the tracker and estimates the categories to which they belong, such as cars, pedestrians, cyclists, etc.
Through the discrimination of the category of the entire track, all the samples on the track actually become learnable to the \emph{Learned Detector}.

%%%%%%%%%%%%%%%%%%%%%%%%%%%%%%%%%%%%%%%%%%%%%%%%%%
\section{Efficient Online Transfer Learning}
\label{sec:case_study}
\begin{figure*}[t]
  \centering
  \includegraphics[width=0.9\textwidth]{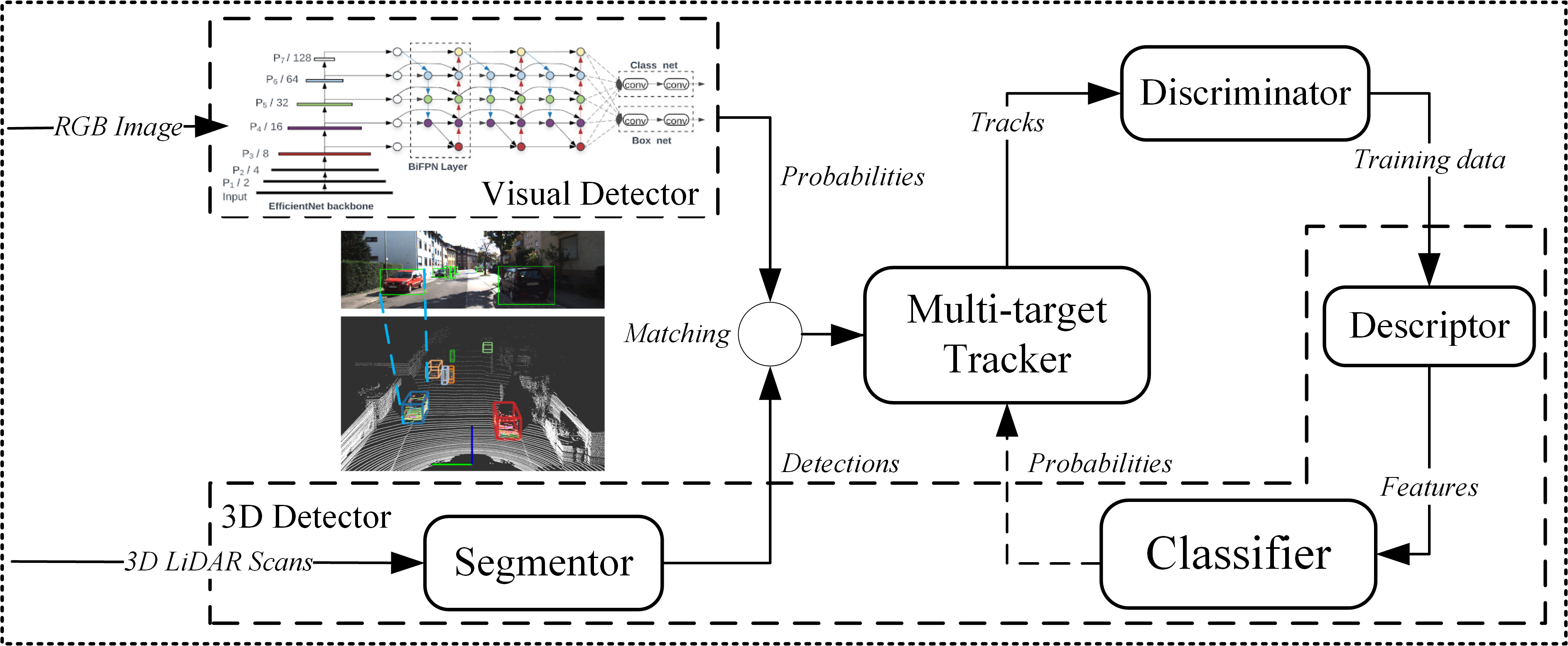}
  \caption{The detailed module diagram of multi-modal-based efficient online transfer learning system for object classification in autonomous driving.}
  \label{fig:implementation}
\end{figure*}

%%%%%%%%%%%%%%%%%%%%%%%%%%%%%%%%%%%%%%%%%%%%%%%%%%
%———————————————— Version - 0.1 —————————————————%
In this section, we present our novel implementation as a specific instantiation of the general framework, with which we can test our system in autonomous driving.
The detailed module diagram of our implementation is shown in Fig.~\ref{fig:implementation}.
Specifically, the RGB image provided by the color camera is input to a pre-trained visual detector based on the state-of-the-art deep learning method~\cite{tan2020efficientdet} to obtain high-confidence label information of the sample.
At the same time, the point cloud generated by the 3D LiDAR is segmented based on the Euclidean clustering algorithm~\cite{RusuDissertation} to form different clusters.
Then, through the known extrinsic calibration parameters between the two sensors and the time stamp synchronization of different sensory data, the 3D bounding box information of the clusters is projected into the 2D space of the synchronized image frame, in order to match the visual detection to obtain the cluster labels and their corresponding probabilities.
It is worth noting that the annotation result from this step is not the final label of the sample used for online learning.
Instead, the clustering results and their annotations are first sent to the multi-target tracker to generate timely the tracks of the object, and then the discriminator will take the track and calculate the probability that the entire track belongs to a certain type of object, so that the final label of all samples on the track for online learning of the 3D LiDAR-based object classifier can be obtained.

Furthermore, our system integrates Autoware's~\cite{autoware} point cloud segmentation and multi-target tracking modules, that are based on traditional methods including Euclidean distance and Kalman filter rather than deep learning, because the former is more in line with Autoware's pursuit of industrial applications, especially for the purposes of trusted, explainable and certifiable.
This idea is also one of the reasons why we are more willing to integrate traditional learning methods like SVM~\cite{yz18iros} and RF (in this paper) into our online framework.
The details of each functional module are given below.

\subsection{Visual Detector}
%———————————————— Version - 0.1 —————————————————%
The visual detector provides positional information of detected objects in 2D images, including the 2D bounding box coordinates, category and prediction score.
It actually plays the role of ``annotator'' in our system, which is mainly used to guide the learning of the classifier in the 3D detector.
In order for the visual detector to provide stable and accurate detection results, we tested three popular deep learning-based detection models including YOLOv3~\cite{redmon2018yolov3}, Faster R-CNN~\cite{ren2015faster} and EfficientDet~\cite{tan2020efficientdet} on the KITTI dataset~\cite{Geiger2012CVPR}.

Table~\ref{tab:detection} summarizes the mAP (mean average precision) of different categories by calculating the arithmetic mean of the results of the three methods at different detection difficulties (i.e. moderate, easy and hard according to KITTI's benchmarks) and the corresponding runtime.
It can be seem that Faster R-CNN and EfficientDet have competitive mAP, and both are better than YOLOv3.
In terms of runtime, YOLOv3 achieved real-time performance while Faster R-CNN has a slower execution speed, meaning that the latter is currently not adaptive for autonomous driving scenarios.
We finally chose EfficientDet-D1 to balance accuracy and real-time requirements, and filter out the output below a predetermined probability threshold\footnote{The threshold was set to 0.5 in our experiments.} to prevent false detections.

\begin{table}[t]
  \caption{Performance evaluation of three deep-learning-based visual detectors on the KITTI dataset}
  \label{tab:detection}
  \begin{center}
    \begin{tabular}{|l|l|l|l|l|}
      \hline
      \textbf{Model} & \textbf{Car} & \textbf{Pedestrian} & \textbf{Cyclist} & \textbf{Runtime}\\
      \hline\hline
      YOLOv3~\cite{redmon2018yolov3} & 59.73$\%$ & 38.51$\%$ & 31.92$\%$ & 21ms\\
      \hline
      Faster R-CNN~\cite{ren2015faster} & 81.57$\%$ & 60.53$\%$ & 57.83$\%$ & 2635ms\\
      \hline
      EfficientDet-D1~\cite{tan2020efficientdet} & 75.83$\%$ & 51.69$\%$ & 46.92$\%$ & 34ms\\
      \hline
    \end{tabular}
  \end{center}
\end{table}

\subsection{3D Detector}
%———————————————— Version - 0.1 —————————————————%
The 3D detector consists of three parts including a segmentor, a descriptor and a classifier.
The 3D LiDAR scan is first segmented into different clusters.
These unlabelled clusters will not be directly input into the descriptor for feature extraction, but will be further processed by the multi-target tracker and discriminator before finally being used for online classifier training.

\subsubsection{Segmentor}
%———————————————— Version - 0.1 —————————————————%
As input of this module, a 3D LiDAR scan is defined as a set of $I$ points (i.e. a 3D point cloud):
\begin{equation}
  P = \{p_i | p_i = (x_i,y_i,z_i) \in \mathbb{R}^3, i = 1, \ldots, I\}
\end{equation}
For better real-time performance (at the expense of accuracy), clusters are extracted from its 3D projection $P' = \lambda P$ where $\lambda = (1,1,0)$, based on the Euclidean distance~\cite{RusuDissertation} between points in 2D space.
A cluster can then be defined as follows:
\begin{equation}
  C_j \subset P', j = 1, \dots, J
\end{equation}
where $J$ is the total number of clusters.
A condition to avoid overlapping clusters is that they should not contain the same points, that is:
\begin{equation}
  \label{eq:clusters}
  C_j \cap C_k = \emptyset, \text{ for } j \neq k, \text{ if } min \| p_j - p_k \|_2 \geq d %Euclidean norm
\end{equation}
where the sets of points $p_j, p_k \in P'$ belong to the point clusters $C_j$ and $C_k$ respectively, if the minimum distance between $p_j$ and $p_k$ is greater than or equal to a given distance threshold $d$.

Before inputting the clusters as objectness detection to the tracker, two more processes are required, namely filtering and annotation.
The former filters out clusters that are too large or too small, which means more attention will be paid to the objects that are more likely to be road participants.
Using a predefined volumetric filter is effective, and can exclude most negative samples and background objects in our case:
\begin{equation}
  \label{eq:volume_filter}
    \overline{C} = \{C_j | w_j \in W,~d_j \in D,~h_j \in H\}
\end{equation}
{where $w_j$ resp. $d_j$ resp. $h_j$ represent the length resp. width resp. height (in meters) of the minimal cube\footnote{The value range of $C_j$ should be adjusted according to the task. In our experiments, $W$ = [0.1, 5], $D$ = [0.1, 5], $H$ = [0.3, 5].} containing $C_j$.
On this basis, the efficiency of cluster-image matching and target tracking is also significantly improved.

Regarding the annotation, it is necessary to match the clustering results with the image detection provided by the visual detector in order to obtain a pre-label of the clusters.
Specifically, we project the 3D bounding box of the clusters onto the 2D image, and determine whether they are the same object by measuring the IoU (Intersection over Union)\footnote{In our experiments, we required an IoU of 0.7 for cars, and an IoU of 0.5 for pedestrians and cyclists, as per KITTI.}.

\subsubsection{Descriptor}
%———————————————— Version - 0.1 —————————————————%
Six features (illustrated in Table~\ref{tab:features}) are extracted from the clusters by the descriptor for later classifier training.
The set of feature values of each sample $C_j$ forms a vector $f_j = (f_1, \ldots, f_6)$.
Features from $f_1$ to $f_4$ were introduced by~\cite{navarro-serment09fsr}, while features $f_5$ and $f_6$ were proposed by~\cite{kidono11iv}.
The selection of these features is mainly based on the needs of real-time perception and lightweight on-board computing for autonomous driving.
For the performance comparison of each feature, please refer to~\cite{kidono11iv,yz19auro,navarro-serment09fsr}.

\begin{table}[t]
  \caption{Features for classifier training}
  \label{tab:features}
  \begin{center}
    \begin{tabular}{|l|l|l|}
      \hline
      \textbf{Feature} & \textbf{Description} & \textbf{Dimension}\\
      \hline\hline
      $f_1$ & Number of points included in the cluster & 1\\
      \hline
      $f_2$ & Minimum cluster distance from the sensor & 1\\
      \hline
      $f_3$ & 3D covariance matrix of the cluster & 6\\
      \hline
      $f_4$ & Normalized moment of inertia tensor & 6\\
      \hline
      $f_5$ & Slice feature for the cluster & 20\\
      \hline
      $f_6$ & Reflection intensity distribution & 27\\
      \hline
    \end{tabular}
  \end{center}
\end{table}

\subsubsection{Classifier}
%———————————————— Version - 0.1 —————————————————%
RF is an ensemble learning method for classification, regression and other tasks that operate by constructing a multitude of decision trees. 
Compared with the SVM used in our previous work~\cite{yz18iros}, RF has inherent advantages of speed and multi-classification.
It also natively supports incremental learning and is conducive to the preservation of knowledge.
The classic RF algorithm uses batch training and generates a series of decision trees based on global data, which cannot be straightforwardly applied to online situations.
A proven online RF method (ORF)~\cite{saffari2009line} referencing the idea of online-bagging, makes full use of the fast and strong generalization capabilities of RFs, has been adopted in our framework.
Later experiments in this paper will show that with the increase in the number of learning samples, the performance of online learned classifier will gradually approach the offline learned one.

Specifically, the statistics in ORF are gathered over time, and the decision when to split on a tree depends on $1)$ the minimum number of samples ($\alpha$) of a node need to reach before splitting, and $2)$ the minimum gain ($\beta$) a split has to achieve.
Thus, a node splits when $|R_j| > \alpha$ and $\exists s \in S : \triangle L(R_j, s) > \beta$, where $\triangle L(R_j, s)$ denotes the gain with respect to a test $s$ in node $j$ which can be measured as:
\begin{equation}
  \label{eq:gain}
    \triangle L(R_j, s) = L(R_j) - \frac{|R_{jls}|}{|R_j|}L(R_{jls})- \frac{|R_{jrs}|}{|R_j|}L(R_{jrs})
\end{equation}
where $R_{jls}$ and $R_{jrs}$ are respectively the left and right partitions made by the test $s$, $|...|$ denotes the number of samples in each partition, and $\triangle L(R_j, s) \geq 0$.
Tests with higher gains will lead to better data splitting in terms of reducing node impurities.
Therefore, when splitting a node, the test with the highest gain is selected as the main decision for the node.

On the basis of the original algorithm, we added support for data streams and improved the functions of small batch sample learning and real-time model storage.
The improved version supports incremental learning with a single or a small number of samples.
For more details, please refer to our released code.

\subsection{Multi-target Tracker}
%———————————————— Version - 0.1 —————————————————%
The multi-target tracker extracted from Autoware is based on the IMM-UKF-PDA framework~\cite{arya20173d,schreier2017bayesian}.
Data association and state estimation are two key issues in target tracking~\cite{nb18eor}, which are even more challenging in autonomous driving.
The idea of the PDA (Probabilistic Data Association) algorithm is that the update state of the target is the weighted sum of the probabilities of each possible target state update.
UKF (Unscented Kalman Filter) is a combination of Unscented Transform (UT) and standard Kalman filter.
Through the UT transformation, the nonlinear system equation can be adapted to the standard Kalman filter system under the linear assumption.
This can effectively overcome the problems of low estimation accuracy and poor stability, therefore provide robust tracking performance.

The IMM (Interactive Multiple Model) algorithm allows to efficiently manage multiple filter models.
Assume that the state equation and measurement equation of the target are as follows:
\begin{equation} \label{eq:state_equation}
  \begin{cases}
    x_{k+1} & = f_{x_k,m_k}+\omega_{k,m_k}\\
    z_k & = h_{x_k,m_k}+\upsilon_{k,m_k}
  \end{cases}
\end{equation}
where $m_k$ is the effective mode at sampling time $k$.
Let the system model set be $M = (m_1, m_2, \dots, m_n)$, where the conversion process conforms to the Markov chain.
Under unconditional constraints, the transition probability $P(m_i(k+1) | m_j(k))$ from $m_j(k)$ to $m_i(k+1)$ is denoted as $\pi_{ji}$.
The probability that model $m_i$ is a matching model at time $k$ is called model probability $P(m_i(k) | z_k)$, which is denoted as $\mu_i(k)$, where $Z = (z_1, z_2, \dots, z_k)$ represents the measurement set of the system.
The initial Markov transition probability satisfies the condition:
\begin{equation} \label{eq:initial_markov}
  \begin{cases}
    \pi_{ji} =P(m_i(k) | m_j(k-1)), m_i,m_j \in M\\
    \sum_{i=1}^{n}\pi_{ji} =1, j = 1, \ldots, n
  \end{cases}
\end{equation}
and the model prediction probability is defined as:
\begin{equation} \label{eq:prediction_probability}
    \mu_{j | i}(k-1) = \frac{\pi_{ji} \mu_j (k-1)}{\sum_{j=1}^{n} \pi_{ji} \mu_j (k-1)}
\end{equation}
Overall, the IMM-UKF filtering process includes input interaction, UKF filtering, model probability update and output fusion. 
For more details, please refer to~\cite{arya20173d,schreier2017bayesian}.

\subsection{Discriminator}
%———————————————— Version - 0.1 —————————————————%
The discriminator generates training samples with label by measuring the \emph{track probability}~\cite{yz18iros} based on Bayes' theorem.
The idea is to measure the likelihood that a track belongs to a specified object such as cars, pedestrians, cyclists, etc., which is defined as follows.
Let $P(y_i|x_i,d_j)$ denote the probability that example $x_i$ is an object with its category label $y_i$ predicted by detector $d_j \in \mathcal{D}$ at the precise time $t$, then the track probability $P(Y_T|X_T,\mathcal{D})$ is computed by integrating the predictions of the different detectors according to the following formula:
\begin{equation}
  \label{eq:px}
  \small
  P(Y_T|X_T,\mathcal{D}) = \frac{odds_{X_{T}}}{1+odds_{X_{T}}}
\end{equation}
where
\begin{equation}
  \label{eq:odds_prod}
  \small
  odds_{X_{T}} = \prod_{i=1}^{t} \prod_{j=1}^{K} odds_{x_{i}}^j
\end{equation}
and
\begin{equation}
  \label{eq:odds}
  \small
  odds_{x_{i}}^j = \frac{P(y_i|x_i,d_j)}{1-P(y_i|x_i,d_j)}
\end{equation}
By thresholding\footnote{The threshold was set to 0.7 in our experiments.} $P(Y_T|X_T,\mathcal{D})$, high-confidence samples are then sent to the descriptor for feature extraction.

%%%%%%%%%%%%%%%%%%%%%%%%%%%%%%%%%%%%%%%%%%%%%%%%%%
\section{EVALUATION}
\label{sec:evaluation}
%%%%%%%%%%%%%%%%%%%%%%%%%%%%%%%%%%%%%%%%%%%%%%%%%%
%———————————————— Version - 0.1 —————————————————%
In this section, we first illustrate our experimental setup, then evaluate individually the performance of the ORF module followed by the entire system, with further analysis of the system architecture through ablation experiments, and finally give qualitative results and discuss the advantages and limitations of our proposed system.

\subsection{Experimental Setup}
%———————————————— Version - 1.0 —————————————————%
The experiments were conducted on the KITTI dataset~\cite{Geiger2012CVPR}, which was collected in urban environments with a car equipped with various sensors including several color and gray-scale cameras and a Velodyne 64-layer LiDAR.
On the one hand, the 7481 training frames (with ground-truth annotations) provided in the 3D object detection benchmark were split into the ratio of 6:2:2 for training, validation and testing respectively.
The image data was used to evaluate EfficientDet and other deep learning based methods (c.f. Table~\ref{tab:detection}) in 2D image detection by 5-fold cross validation, while the point cloud data was used to evaluate the performance of ORF-based classifier (c.f. Sec.~\ref{sec:orf_evaluation}).

On the other hand,  considering the significance of temporally continuous scenes (i.e. time-adjacent frames) to the online learning framework, the raw data (five categories including city, residential, road, campus, and person) provided by KITTI were used to online train the 3D LiDAR-based classifier (c.f. Sec.~\ref{sec:system_evaluation}), which does not contain any annotations in the image or point cloud.
Additionally, scene fragments that last longer than 60 seconds were excluded, since most of them are used for visual odometry task, which means that the scene is typically monotonous and the objects contained are usually stationary vehicles.
Moreover, an under-sampling was applied to process training samples in order to reduce the negative effect of unbalanced distribution of learning samples (especially with many cars and few cyclists) on the performance of the learned classifier.

Our framework has been fully implemented into ROS with very high modularity.
All components are ready for download and used by the community.
All the experiments reported in this paper were performed with Ubuntu 18.04 LTS (64-bit) and ROS Melodic, with an Intel i9-9900K CPU and 64GB RAM, and a NVIDIA GeForce RTX 2080 GPU with 8GB RAM.

\subsection{ORF Performance}
\label{sec:orf_evaluation}
%———————————————— Version - 1.0 —————————————————%
In order to evaluate the performance of the ORF module, we randomly selected 1000 samples from our training subset to form a data stream as the input of ORF.
Specifically, every time the classifier learns 100 samples, we will save a model locally and evaluate it on our test subset to report as the result of an iteration.
We report the results of the first ten iterations.
In addition, a classic batch RF-based classifier~\cite{scikit-learn} offline trained with the same 1000 samples is served as the baseline for comparison.
Regarding the structure of the forest, both online and offline manners are consistent in parameter settings, i.e., $trees = 100, depth = 50, epochs = 20, split\_threshold = 50$, while the other parameters remain the default values as in the released code.

The experimental results are shown in Fig.~\ref{fig:online_vs_offline}.
F1-score is used as the evaluation metric, including the average accuracy (ACC, numerically equal to micro-F1-score) and the macro average (MaA, i.e. macro-F1-score).
It can be seen that as the number of training samples increases, the performance of the online learned classifier quickly approaches the offline trained one.

\begin{figure}[t]
  \centering
  \includegraphics[width=\columnwidth]{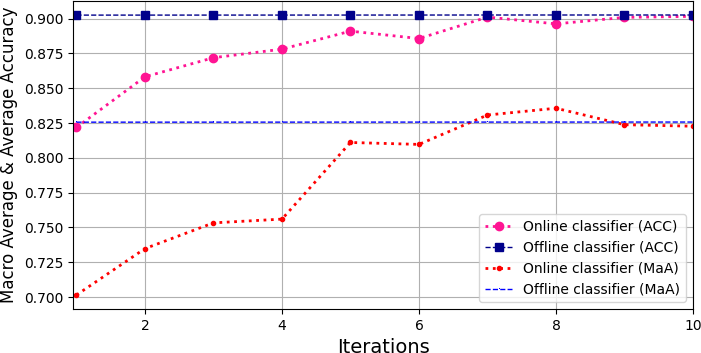}
  \caption{Performance comparison of the ORF-based online learned and the RF-based offline trained 3D LiDAR object classifiers.}
  \label{fig:online_vs_offline}
\end{figure}

\subsection{System Performance}
\label{sec:system_evaluation}
%———————————————— Version - 1.0 —————————————————%
In order to evaluate the performance of our proposed system, we first compare the classification results of the 3D LiDAR-based classifier learned online with KITTI's raw data under our framework (c.f. Fig.~\ref{fig:implementation}), with the independently learned (with ground-truth annotations) ORF-based classifier reported in the previous section.
We report the results after the final iteration (i.e. after learning 1000 samples), as shown in Fig.~\ref{fig:classification_results}.
Two confusion matrices containing three categories (i.e. car, pedestrian and cyclist) are used to evaluate the performance of the two classifiers.
The abscissa indicates the predicted result while the ordinate represents the true label.
Different color depths correspond to proportions, and the darker the color, the higher the percentage of correct classification.

\begin{figure}[t]
  \centering
  \includegraphics[width=\columnwidth]{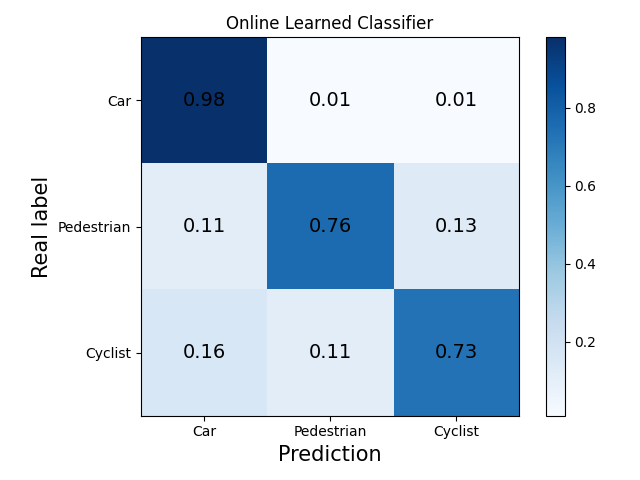}\\
  \vspace{0.2cm}
  \includegraphics[width=\columnwidth]{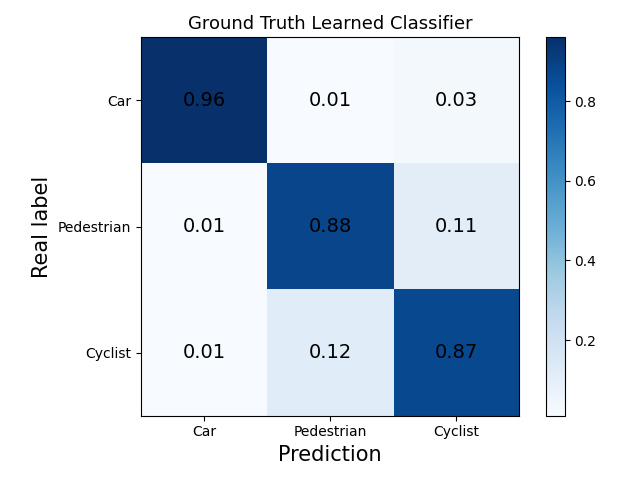}
  \caption{Confusion matrices of online learned 3D LiDAR-based classifiers respectively under our proposed system (with raw data) and using ORF alone (with ground-truth annotations).}
  \label{fig:classification_results}
\end{figure}

The results show that although the performance of the 3D LiDAR-based classifier learned under our proposed system cannot surpass the ORF-only classifier with limited samples, there is no huge gap between the two classifiers.
Since the proportion of vehicles in the test subset is the highest, it is reasonable that fewer vehicles are incorrectly classified into other categories.
There are relatively more incorrect classifications of pedestrians and cyclists because the segmentor module in our system suffers from difficulties of segmenting people or bicycle groups, so that the classifier learns false positive samples.

\begin{figure}[t]
  \centering
  \includegraphics[width=\columnwidth]{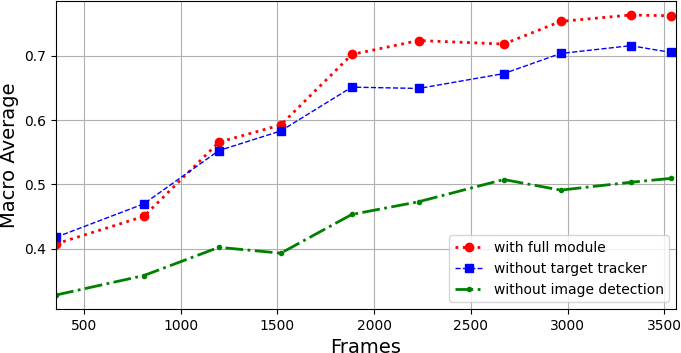}
  \caption{Results of ablation experiments.}
  \label{fig:ablation}
\end{figure}

Furthermore, we analyze and verify our system architecture through ablation experiments.
It can be seen from Fig.~\ref{fig:ablation} that the performance of the classifier learned without the visual detector but only relying on a volumetric template cannot be compared with other results.
This is because using only the size of the cluster to determine its category produces considerable number of false positive samples.
The classifier under the system without the target tracker learns better in simple scenarios (i.e. the first 1500 frames), but the learning efficiency decreases as the scenarios become more complex (i.e. more pedestrians and cyclists appeared in 1500-2000 frames and 2500-3000 frames).
Instead, the classifier learned under the complete system shows the best performance.

\subsection{Qualitative Analysis}
%———————————————— Version - 1.0 —————————————————%
We mainly focus on the qualitative analysis of three aspects, as shown in Fig.~\ref{fig:qualitative_analysis}.
The first is about the segmentation of the point cloud.
Overall, the segmentor shows good performance in both simple and complex scenes, but when facing objects that are too close (forming a group), there is still room for improvement.
For example, in the area shown in the red box in the lower part of Fig.~\ref{fig:qualitative_analysis}, people and benches are clustered as a whole.
The second is the performance of the visual detector that plays the role of the trainer.
We can see that image detection still shows acceptable performance in complex environments, and the matching with point cloud clusters is also robust.
The third point is about the role of the multi-target tracker.
It can be seen from the upper part of Fig.~\ref{fig:qualitative_analysis} that although the cyclist has left the camera's visual range, the classifier can still learn the sample of the object in the point cloud through the association of the tracker (indicated by the blue box).
Similar situations are shown in the lower part of Fig.~\ref{fig:qualitative_analysis}.
This undoubtedly greatly improves the learning efficiency of the entire system.

\begin{figure}[!b]
  \centering
  \includegraphics[width=\columnwidth]{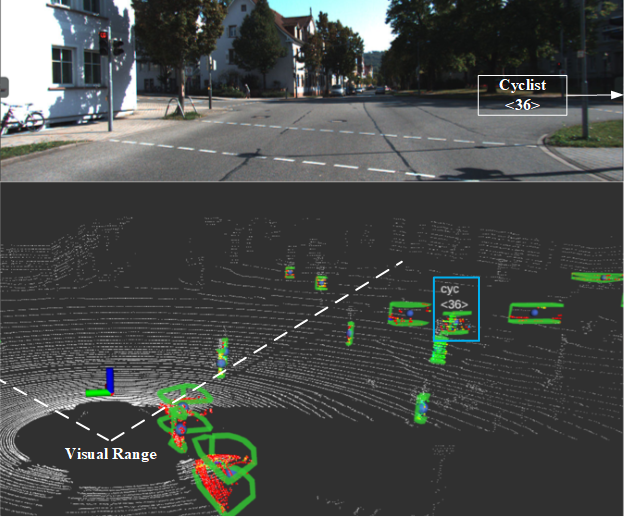}\\
  \vspace{0.2cm}
  \includegraphics[width=\columnwidth]{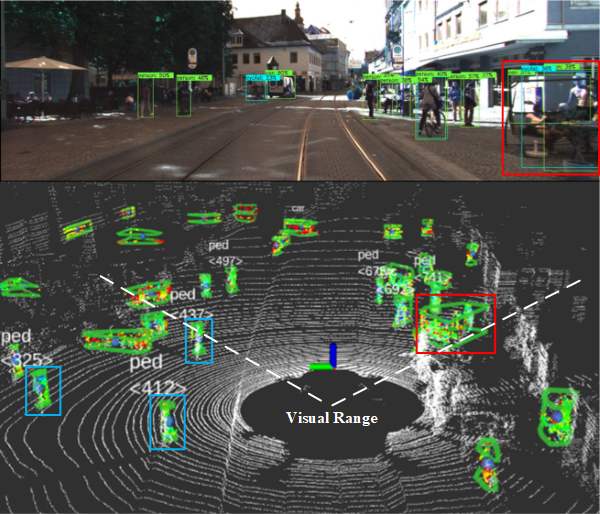}
  \caption{Two pairs of synchronized frames include the color image and its corresponding 3D LiDAR scan. The coordinate axis in the point cloud represents the position of the 3D LIDAR sensor. The colored point cloud enclosed by the bounding box represents the samples input to the classifier. Note that only those labelled samples will be learned by the classifier, such as the cyclist in the upper image. The upper and lower parts respectively show the performance of our system in simple and complex scenarios.}
  \label{fig:qualitative_analysis}
\end{figure}

%%%%%%%%%%%%%%%%%%%%%%%%%%%%%%%%%%%%%%%%%%%%%%%%%%
\section{CONCLUSION}
%%%%%%%%%%%%%%%%%%%%%%%%%%%%%%%%%%%%%%%%%%%%%%%%%%
%———————————————— Version - 1.0 —————————————————%
In this paper, we presented a novel implementation of the previously proposed multi-sensor online transfer learning framework, which is capable of efficiently learning a 3D LiDAR-based multi-class classifier on-the-fly from a monocular camera-based detector in urban environments.
Specifically, we explored how to use today's popular deep learning technology to help learn from data that is not easy to interpret, such as sparse point clouds.
To do so, leveraging a powerful multi-target tracker provided by Autoware, we actually established a pipeline between the pre-trained detector and the learning detector to achieve knowledge transfer, and used ORF to realize rapid incremental learning. 
A very promising feature of the proposed solution is that a new object classifier can be learned and updated directly from the deployment environment, thus getting rid of the dependence of model learning on manual data annotation on the one hand, and improving the system's ability to adapt to environmental changes on the other hand.

For the first time, we deployed our online learning framework in the field of autonomous driving and demonstrated promising results through experiments.
The proposed system has been fully implemented into ROS with a high level of modularity, and open sourced to the community.

Despite the encouraging results, there are still several aspects that can be improved.
Future work will include improving the quality of online learning samples thus the classifier performance, and experimenting with some other datasets (for different environments) as well as our own self-driving cars equipped with low-power computing units.
% Rui, only for you, "improving the quality of online learning samples" means A) clustering, B) system loop closure ;)

%%%%%%%%%%%%%%%%%%%%%%%%%%%%%%%%%%%%%%%%%%%%%%%%%%
\section*{ACKNOWLEDGMENT}
%%%%%%%%%%%%%%%%%%%%%%%%%%%%%%%%%%%%%%%%%%%%%%%%%%
%———————————————— Version - 1.0 —————————————————%
We thank the Autoware Foundation and Dr. Tixiao Shan for his initial barebone tracker package. UTBM is a member of the Autoware Foundation.

\bibliographystyle{IEEEtran}
\bibliography{rui21itsc}
\end{document}